\documentclass[10pt, a4paper]{article}

\usepackage{lrec-coling2024} % this is the new style
\usepackage{enumitem}
\usepackage{booktabs}
\usepackage{float}
\usepackage{makecell} 

\usepackage{tabularx, booktabs}
\usepackage{colortbl}
\usepackage{graphicx}

\title{Lightweight Connective Detection Using Gradient Boosting\\ \vspace*{.5\baselineskip} 
}

\name{Mustafa Erolcan Er$^1$, Murathan Kurfalı$^2$, Deniz Zeyrek$^1$} 

\address{$^1$Cognitive Science Dept., Graduate School of Informatics, Middle East Technical University\\
$^2$Sensory-Cognitive Interaction Lab, Department of Psychology, Stockholm University \\
         erolcan@metu.edu.tr, murathan.kurfali@su.se, dezeyrek@metu.edu.tr\\}

\abstract{
In this work, we introduce a lightweight discourse connective detection system. Employing gradient boosting trained on straightforward, low-complexity features, this proposed approach sidesteps the computational demands of the current approaches that rely on deep neural networks. Considering its simplicity, our approach achieves competitive results while offering significant gains in terms of time even on CPU. Furthermore, the stable performance across two unrelated languages suggests the robustness of our system in the multilingual scenario. The model is designed to support the annotation of discourse relations, particularly in scenarios with limited resources, while minimizing performance loss.   
\\ \newline  }
\begin{document}

\maketitleabstract

\section{Introduction}

% A bottom-up systematic study of natural language begins with examining how morphemes interact with suffixes and prefixes at the word level. This area of linguistics is called morphology. In the next stage, sentences are formed as a result of words interacting with each other according to certain rules. Whether these sentences form a whole in accordance with natural language rules is the subject of syntax. When we take this systematic one step further, we encounter structures called discourse formed by sentences. 
% Discourse might be considered as a phenomenon that covers the natural language at the document level by placing it above the word level and sentence level as their composition.

Recent advancements in deep learning have significantly improved state-of-the-art performances in natural language processing (NLP), and discourse parsing is no exception. Yet, despite these performance gains, these models demand high computing resources, which greatly hinders their usability, as many researchers around the world still lack access. Moreover, these models often act as black-box solutions, without providing any linguistic/theoretical insights regarding the task at hand. In our current submission, we present a lightweight detection system for connectives, which are considered as one of the most important building blocks of discourse structure.

 Among various approaches to discourse structure, such as RST \cite{mann1987rhetorical} and SDRT \cite{lascarides2007segmented}, PDTB \cite{prasad2014reflections} remains the largest annotated dataset \cite{prasad2014reflections} involving discourse-level annotations. 
PDTB adopts a connective-based approach, where connectives are the anchors of discourse relations that hold between two text spans that have an abstract object interpretation, such as propositions or eventualities \cite{prasad2014reflections}. 
% The challenge is that most connectives have both discourse (DC) and non-discourse usage (NDC) based on the contexts in which they appear. 
The challenge lies in distinguishing between connectives that function as discourse connectives (DC) and those that do not, known as non-discourse connective (NDC) usage.
Consider examples (\ref{ex:paris}) and (\ref{ex:eng-fr}): 
\begin{enumerate}
    \item \label{ex:paris} He went to Paris for a vacation and visited the famous Eiffel Tower.
\item \label{ex:eng-fr} He speaks English and French. 
\end{enumerate}
\vspace{-0.2cm}\hspace{2cm}  (from \cite{bacsibuyuk2023usage})
    
PDTB recognizes the \textit{and} in the first example as a discourse connective whereas, in the second example, it does not, as it simply links two noun phrases. Thus, the first step in the PDTB annotation process is the detection of the connectives with discourse usage in a given text piece.  
In the current work, we address this issue using a lightweight model that utilizes linguistic features to efficiently identify discourse connectives without the need for specialized hardware, such as GPUs, which are still not available to most researchers worldwide. We train and evaluate our model in two languages, English (PDTB 2.0) and Turkish (Turkish Discourse Bank (TDB) 1.0 \cite{zeyrek2013turkish}). The contributions of our work are: 

\begin{enumerate}
    \item We introduce a fast machine-learning model that detects connectives.
    \item We show that this model achieves results close to state-of-the-art models.
    \item We argue that verb-based features are the most important aspects of our lightweight connective detection model.
\end{enumerate}

The paper is structured as follows. In Section \ref{sec:related-works}, we introduce two lines of research that deal with connective detection
and briefly summarize recently developed discourse parsers that are shown to work in Turkish as well as English. Section \ref{sec:method} introduces our method, and
Section \ref{sec:exp-setting} the experimental setting as well as the data and baselines. In Section \ref{sec:evaluation} we evaluate our model, and finally, in Section \ref{sec:conc} we draw some conclusions. 

% make them impractical in resource-limited scenarios

\section{Related Work}
\label{sec:related-works}
%murathan:
Reflecting the overall trend in the field, the literature on discourse parsing can be roughly divided into two parts: the body of works before, and after the emergence of neural networks (NNs). Before the solutions based on neural networks became the default approach, the methods relied more on traditional approaches such as feature engineering or annotation projection \cite{wellner2007automatically,pitler2009using,versley2010discovery}.

\begin{table*}[htbp]
    \centering
    \small
    \begin{tabular}{lccccc}
        \toprule
        \textbf{Model} & \textbf{Learning Rate} & \textbf{Max Depth} & \textbf{N Estimators} & \textbf{Max Delta Step} & \textbf{Min Child Weights} \\
        \midrule
        PDTB 2.0  & 0.2 & 8 & 500 & 4 & 1 \\
        PDTB 2.0 (Weighted)  & 0.30 & 8 & 400 & 4 & 1 \\
        TDB 1.0  & 0.15 & 10 & 500 & 4 & 1 \\
        TDB 1.0 (Weighted) & 0.15 & 8 & 400 & 4 & 1 \\
        \bottomrule
    \end{tabular}
    \caption{Best Parameters for PDTB 2.0 and TDB 1.0 Datasets. Weighted refers to the classifiers trained with the "weighted" loss.}
    \label{tab:best_parameters}
\end{table*}

Following the deep learning revolution, led by the increase in the available computing power and the amount of data, NN-based solutions slowly replaced linguistic features, and more black-box approaches have become popular \cite{hooda2017argument,kurfali2020labeling,kutlu2023toward}. Most prominently, the recent DISRPT 2021 \cite{zeldes2021disrpt} and 2023 \cite{braud2023disrpt} shared tasks have received only transformer \cite{vaswani2017attention}-based solutions to a range of languages including English and Turkish \cite[e.g.,][]{gessler2021discodisco,metheniti2023discut,anuranjana2023discoflan}, with the exception of the TMVM model by \citet{donicke2021delexicalised}, which utilized linguistic features derived from syntactic trees. \citet{gessler2021discodisco} also stands out by integrating linguistics features into transformers.

\section{Approach}\label{sec:method}

The proposed connective detection model takes raw natural language data as input and determines which tokens are connectives. The task is modeled as a three-way token classification task, where each token can belong to one of three categories:
\begin{itemize}
    \item \textit{O}: The token is not part of a connective span.
    \item \textit{B-Conn}: The token marks the beginning of a connective span. It can represent the entire span of the connective, as in single-word examples like \textit{because}, or the first word of a phrasal connective, such as \textit{on} in \textit{on the other hand}.
    \item I-Conn: The token is the second or a subsequent word in a phrasal connective, e.g., \textit{other} in \textit{on the other hand}.
\end{itemize}

A computationally cheap and fast explicit connective detection algorithm should use symbolic or traditional ML-based approaches instead of deep learning architectures. At the same time, the features used by ML-based algorithms should be produced by algorithms with a time complexity lower than the inference time complexity of the ML model. For this purpose, we preferred to use gradient boosting to train our model. Gradient boosting is an ensemble method determining the optimal predictive model to enable us to use the decision trees more effectively \cite{friedman2001greedy}. 

This iterative algorithm starts with a naive prediction (mostly an average line) to capture the target values. In the second iteration, the residual between this prior prediction and the observed targets is calculated and a decision branch is adapted to decrease the sum of residuals. Repeating this process until the sum of residuals is minimized gives us a final decision tree for our classification task. We use the XGBoost \cite{chen2015xgboost} library to implement gradient boosting on our datasets. 
% Given that the features used in our algorithm should be extracted with a low time complexity, 

We 
% investigated potential features and 
decided to incorporate three groups of features to our model.
% We can classify the features we use in training our model into three groups. 
The first group involves verb-based features. These are the main features for our model and involve:  
% because they are easy to detect and highly correlated with the occurrence of discourse connectives. 
% The features in this group can be listed as below:

\begin{itemize}
    \item Whether any of the three words before and three words after a candidate token is verb or not.
    \item Whether the current word is verb or not.
    \item The token-based distance of the current word to the previous and the following verbs. 
\end{itemize}

The second group of features involves word-based features consisting of features such as the capitalization of words, word length, and a unique ID assigned to each word in the data, all of which can be produced with O(n) time complexity.

The last group of features includes position-based features, by which 
we could produce in O(n) time complexity, too. These involve the position of the current word in the sentence, also including the length of sentences based on words.

\begin{table*}[ht]
\centering
\begin{tabular}{lccccr}
\toprule
Dataset & B-Conn &I-Conn & O & Connective Proportion(\%) \\
\midrule
& \multicolumn{4}{c}{TDB} \\ \midrule
Training & 7,044 & 1,259 & 385,256 & 2.11 \\
Development & 773 & 130 & 45,939 & 1.93 \\
Test & 849 & 165 & 45,944 & 2.16 \\ \midrule
& \multicolumn{4}{c}{PTDB} \\ \midrule
Training & 23,848 & 4,499 & 1,032,851 & 2.67 \\
Development & 953 & 159 & 38,656 & 2.80 \\
Test & 1,245 & 238 & 54164 & 2.67 \\
\bottomrule
\end{tabular}
\caption{The distribution of labels in the datasets. Refer to Section \ref{sec:method} for the label definitions. The last column denotes the proportion of all connectives to the total number of tokens.}
\label{table:dataset_statistics}
\end{table*}

We used the XGBoost library to train our model with gradient boosting. The XGBoost library offers a wide choice of parameters for gradient boosting. Thus, we performed parameter tuning on learning\_rate (contributions of each tree to the final model), max\_depth (maximum depth of each tree), n\_estimators (number of trees generated by the model), max\_delta\_step (a parameter that is useful for imbalanced datasets by preventing the weights from updating too much) and min\_child\_weights (a parameter to control the overfitting problem) which we consider to be the most important ones among these parameters. We used the grid search algorithm \cite{chicco2017ten} to choose the most effective tuning among these three parameters. Grid search systematically runs the different combinations of parameters and uses cross-validation \cite{stone1974cross} to find the best combination based on the performance.  Recognizing the limited size of our dataset, we applied 3-fold cross-validation in our experiments to ensure a balance between model training time and validation robustness.

\begin{table*}[htbp] % Use 'table*' to span the table across two columns
\centering
\begin{tabular}{@{}lllll@{}} 
\toprule
\textbf{Model} & \makecell{\textbf{Precision} \\ \textbf{(\%)}} & \makecell{\textbf{Recall} \\ \textbf{(\%)}} & \makecell{\textbf{f-score} \\ \textbf{(\%)}} & \makecell{\textbf{Inference} \\ \textbf{Time (sec)}} \\
\midrule 

DisCut2023 \cite{metheniti2023discut} & 95.49 & 91.89 & 93.66 & --\\
DiscoDisco \cite{gessler2021discodisco} & 92.93 & 91.15 & 92.02 & -- 
\\
Segformers \cite{bakshi2021transformer} & 89.73 & 92.61 & 91.15 &-- \\
DisCut \cite{ezzabady2021multi} & 93.32 & 88.67 & 90.94 & -- \\
TMVM \cite{donicke2021delexicalised} & 85.98 & 65.54 & 74.38 & --\\  \hline 
BERT Baseline & 92.63 & 91.88 & 92.25 & 3.13 \\ \hline
Our Model & 89.10 & 78.71 & 83.58 & 0.02 (1.33*)                            \\
Our Model (Weighted) & 70.00 & 86.02 & 77.19 & 0.02 (2.03*) \\
\bottomrule
\end{tabular}
\caption{Comparison of the Baseline Models and Our Model over PDTB 2.0 Using DISRPT Data Splits. * denotes inference time on CPU for our lightweight model.}
\label{tab:baseline_models_eng}
\end{table*}

The dataset suffers from severe imbalance as discourse connectives do not occur as often. To deal with this, we also train our models with the weighted loss. We used inverse frequency weighting to determine the label weights. That is, for each   \(i\) in our dataset, we computed  \(w_i\) as
\[w_i = \frac{N}{C \cdot n_i}\]

where  \(N\) is the total number of instances, \(C\) is the number of unique classes and \(n_i\) is the number of instances belonging to class \(i\). 

Weighted loss is a method used in imbalanced data to ensure that minority class data points contributes more to the model. The idea behind weighted loss is to assign a higher weight to the minority class data points while assign a lower weight to the majority class data points when computing the loss. Thanks to this approach, mistakes on the minority class become more "costly" for the model, causing it to pay more attention to correctly classifying instances of the minority class.

The best parameters according to the grid search are provided in Table \ref{tab:best_parameters}.

\section{Experimental setting}\label{sec:exp-setting}

\subsection{Data}\label{sec:data}

In our experiments, we followed the training, development, and test splits proposed in DISRPT 2021 \cite{zeldes2021disrpt} to facilitate direct comparison of our models with the state-of-the-art systems evaluated there. The Turkish data in DISRPT is sourced from TDB 1.0 \cite{zeyrek2013turkish}, while the English data is based on PDTB 2.0 \cite{prasad2008penn}. The distribution of the labels in the respective datasets are provided in Table \ref{table:dataset_statistics}. DISRPT data uses these datasets without any pruning. Thus, our models are trained to explicit discourse connectives including discontinuous connectives such as "if .. then", "either .. or", etc. in addition to continuous or single word connectives. Alternative Lexicalizations (AltLex) connectives are also included in these datasets. AltLexes are not connective on their own but can act as connective when combined as multi word expressions.

\subsection{Baseline Models}\label{sec:baseline}

To put our results into perspective, we compare our model's performance against the best-performing systems in DISPRT 2021 and 2023 shared tasks.
Additionally, we report the performance of a vanilla  BERT model fine-tuned on the training set\footnote{We used the \textit{bert-base-cased} for English and the BERTurk model \cite{schweter2020berturk} for Turkish.} \cite{devlin2018bert}, to represent the current go-to approach for performing this task. We follow the standard token classification procedure using the default parameters and report the average performance across four different runs. The BERT baseline also provides insights into the time efficiency of our model, as that information is not available for the other baselines. It should be noted that all baselines, except for TMVM, are based on deep neural networks.

\begin{table*}[htbp] % Use 'table*' to span the table across two columns
\centering
\begin{tabular}{@{}lllll@{}} 
\toprule
\textbf{Model} & \makecell{\textbf{Precision} \\ \textbf{(\%)}} & \makecell{\textbf{Recall} \\ \textbf{(\%)}} & \makecell{\textbf{f-score} \\ \textbf{(\%)}} & \makecell{\textbf{Inference} \\ \textbf{Time (sec)}} \\
\midrule 

DiscoDisco \cite{gessler2021discodisco} & 93.71 & 94.53 & 94.11 & -- 
\\

DisCut2023 \cite{metheniti2023discut}&92.34& 93.21& 92.77 & -- \\
Segformers\cite{bakshi2021transformer} & 90.42 & 91.17 & 90.79 & -- \\
DisCut \cite{ezzabady2021multi}  & 90.55 & 86.93 & 88.70 & -- \\
TMVM \cite{donicke2021delexicalised} & 80.00 & 24.14 & 37.10 & --\\  \hline 
BERT Baseline & 92.36 & 92.89 & 92.62 & 5.09 \\ \hline
Our Model & 87.41 & 71.96 & 78.94 & 0.01 (1.17*) \\
Our Model (Weighted) & 82.42 & 82.33 & 82.38 & 0.01 (1.55*) \\
\bottomrule
\end{tabular}
\caption{Comparison of the Baseline Models and Our Models over TDB 1.0 Using DISRPT Data Splits. * denotes inference time on CPU for our lightweight model.}
\label{tab:baseline_models_tur}
\end{table*}

\section{Results and Discussion} \label{sec:evaluation}
\subsection{Results}

We evaluated the performance of our model using the official evaluation script of DISRPT 2021.\footnote{https://github.com/disrpt/sharedtask2021} The evaluation criteria are based on exact span matching, meaning that partial detection of phrasal connectives, such as identifying "\textit{because}" within "\textit{That's because}", does not contribute to the overall accuracy. For each language, micro-averaged precision, recall, and F-scores are reported.

The results of our system for English and Turkish are provided in Table \ref{tab:baseline_models_eng} and Table \ref{tab:baseline_models_tur}, respectively. Despite our model's simplicity and reduced complexity, it demonstrates competitive performance when compared against the strong baselines. The best performances achieved in English and Turkish are very close to each other, suggesting that the model is robust across languages with different linguistic characteristics. Moreover, it must be highlighted that our submission outperforms the feature-based baseline, TMVM, in both languages, with the difference in Turkish being almost three-fold. We believe that this finding demonstrates the effectiveness of our set of features and further justifies their applicability to different languages.

Switching to weighted loss led to mixed results. In Turkish, the weighted loss increased the overall performance by 3 points; however, in English, it had a negative effect. Yet, in both cases, weighted loss significantly increased the recall of our models as expected. These findings indicate that while the approach increases the model's ability to identify true positive cases, its impact on precision, hence the overall performance, is language-dependent and requires further investigation.

On the other hand, our models achieved inference speeds at least three times faster than the BERT Baseline, despite being run on a CPU, unlike the BERT model which was trained and evaluated on a GPU. When both models are run on a GPU, the difference becomes nearly 250 times. This confirms that our model is indeed computationally less demanding, making it suitable for scenarios with limited computational resources.

\subsection{Feature Importance}

\begin{figure}[tp] % Using 'figure*' instead of 'figure' to span two columns
    \centering
    \includegraphics[width=0.5\textwidth]{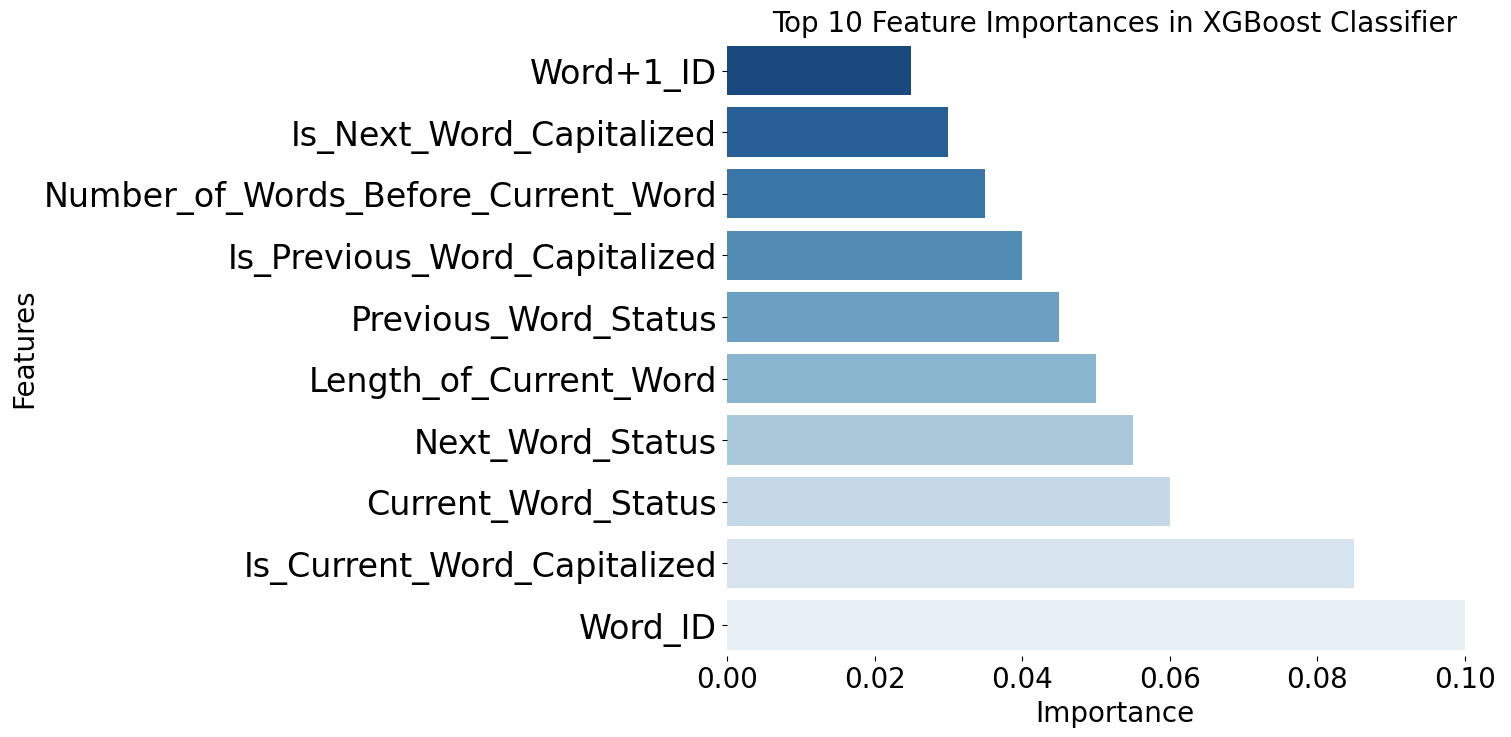} % 'textwidth' spans the figure across the page width
    \caption{Feature importance in PDTB 2.0 for our best model}
    \label{fig:feature-importance-pdtb}
\end{figure}

\begin{figure}[tp] % Using 'figure*' to allow the figure to span two columns
    \centering
    \includegraphics[width=0.45\textwidth]{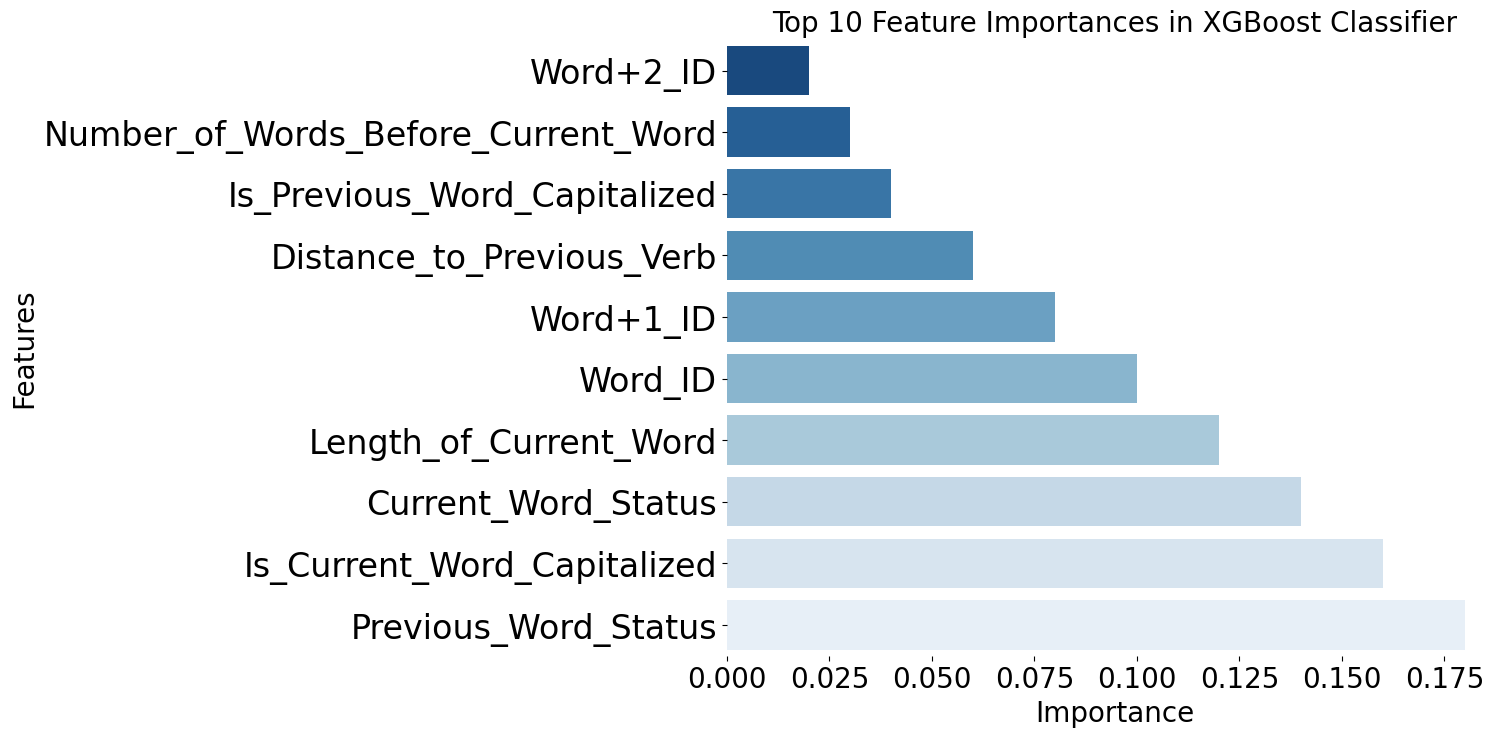} % Adjusting width to textwidth to span the whole page width
    \caption{Feature importance in TDB 1.0 for our best model}
    \label{fig:feature-importance-tdb}
\end{figure}

After training our model, we performed a feature importance test to determine which features made the highest contribution to the detection of  DCs in TDB 1.0 and PDTB 2.0. The most important features detected by our best models in two languages are listed in Figure \ref{fig:feature-importance-pdtb},  Figure \ref{fig:feature-importance-tdb}. 

\begin{table*}[ht]
\centering
\scriptsize
\begin{tabular}{@{}lcccccccccc@{}}
\toprule
\textbf{Connective} & \multicolumn{2}{c}{\textbf{Number of Correct Predictions}} & \multicolumn{2}{c}{\textbf{Number of Incorrect Predictions}}& \textbf{Accuracy (\%)} \\ 
\cmidrule(r){2-3} \cmidrule(r){4-5} 
& True Positive (TP) & True Negative (TN) & False Positive (FP) & False Negative (FN)&  \\
\midrule
and & 204 & 619 & 21 & 40 & 93.10  \\
for & 11 & 403 & 1 & 10 & 97.41  \\
then & 11 & 2 & 2 & 3 & 72.22  \\
Once  & 0 & 0 & 3 & 1 & 0  \\

\toprule
\toprule

ve (and) & 181 & 477 & 33 & 25 & 91.90  \\
için (for) & 90 & 88 & 20 & 2 & 89.00  \\
Sonra (After)  & 15 & 2 & 4 & 2 & 73.92  \\
aksine (contrary to)  & 0 & 1 & 0 & 2 & 33.33  \\
\bottomrule
\end{tabular}
\caption{Error Statistics for Selected Connectives in English (above) and Turkish (below). The top two connectives are the most frequent ones; the bottom two are the most mispredicted that occur at least three times.}
\label{tab:error_table}
\end{table*}

As seen in the figures, word-based features such as Word ID and Capitalization check are prominent for PDTB. For TDB, the most critical feature is the information on whether the previous word is a verb (Previous\_Word\_Status). Additionally, while the status of the current word as a verb (Current\_Word\_Status) significantly contributes to the model for both languages, verb information of the next word for English and the previous word for Turkish stand out. We believe this may be attributed to the differences in word order between Turkish and English.

As shown in \cite{pitler2009using}, constituent tree-based features such as self category, parent category, sibling category provide very successful results in detecting explicit connectives. However, since annotated trees aligned with raw data are needed to derive these features, deriving these features also has an additional annotation cost. In fact, since the annotation process of a dataset with the PDTB formalism is easier than the constituent tree annotation process, deriving the features to be used for automatic annotation may even cause higher costs than handmade annotation. This shows that our system, in addition to being lightweight compared to deep learning models, is also lightweight compared to classical approaches in terms of producing features effectively and at low cost.

\subsection{Error Analysis}
\label{sec:error}
% To see where the errors of our model appears, 
In this section, we discuss our model's performance through error analysis. We present the error distribution for selected connectives in Table \ref{tab:error_table} and discuss some examples. The table highlights the first two connectives as those with the highest occurrence in our dataset, while the last two are identified as the most frequently mispredicted connectives above the specific threshold of 3. For Turkish data, the model tends to over-predict discourse connective (DC) usage over non-discourse connective (NDC) usage while in the PDTB, it is more cautious, often missing instances where connectives serve as DCs.

% whereas, in PDTB, it is more cautious, often missing instances where connectives serve as DCs.

% (tables explanation) 

The examples below are provided to highlight the mistakes of our model. We show the mispredicted tokens by underlining, correctly predicted ones in bold fonts. 

Example (\ref{ex:laiklik}) showcases an unusual case where our model incorrectly identifies a noun in the Turkish dataset, \textit{aklı} ('mind'), as a discourse connective.

\begin{enumerate}[resume]
    \item\label{ex:laiklik} Laiklik zaten, inançlara saygı duyarak \underline{aklı} özgürleştirmektir.(False Positive)\\
    `Secularism already means liberating \underline{the mind} by respecting beliefs.' 
    \end{enumerate}
 This error is noteworthy because the sentence does indeed contain a connective that expresses a manner relation, specifically through the (intra-sentential) suffixal connective -arak attached to the verb preceding \textit{aklı}. Yet, such suffixal connectives are later added to the TDB in its 1.2 version \cite{zeyrek2022description} and are missing in the DISRPT training data. We have spotted several more cases exhibiting the same behavior which suggests that our model is generalizing to the connectives that are not seen in its training data.
 
Examples (\ref{ex:uygun}) and (\ref{ex:Gannett}) illustrate one of the most common mistakes of our model, both in Turkish and English datasets. In Turkish, it includes a phrasal connective \textit{zaman da} ('when' used with the focus particle); yet, our model only identifies the first part, \textit{zaman} ('when'), missing the focus particle, \textit{da}. In English, the system only recognizes \textit{for}, missing the rest of the connective.  Due to the strict evaluation strategy that requires an exact span match, this prediction is classified as misprediction. Overall, the phrasal connectives are particularly challenging.
   
    \begin{enumerate}[resume]
    \item\label{ex:uygun} Uygun düştüğü sanıldığı \textbf{zaman} \underline{da} hemen birbirlerinin üzerinden kayıp gideceklerdi. (False Negative)\\
    `\textbf{When} people thought [it] fits, they would immediately slip over each other'.
    \end{enumerate}

    \begin{enumerate}[resume]
    \item\label{ex:Gannett} \textbf{For} \underline{instance}, Gannett Co. posted an 11\% gain in net income, as total ad pages dropped at USA Today, but advertising revenue rose because of a higher circulation rate base and increased rates. (False Negative)
    % \item\label{ex:newspaper} Newspaper publishers are reporting mixed third-quarter results, \underline{aided} by favorable newsprint prices and hampered by flat or declining advertising linage, especially in the Northeast. (False Positive) 
    % \item\label{ex:courts} Should the courts uphold the validity of this type of defense, ASKO will \textbf{then} ask the court to overturn such a vote-diluting maneuver recently deployed by Koninklijke Ahold NV (False Negative)
\end{enumerate}

% which we will explore in the subsequent studies.
% is an MWE connective. 

\section{Conclusion and Further Studies} \label{sec:conc}

In this study, we introduced a lightweight, gradient-boosting-based system for detecting discourse connectives, achieving competitive performance with significantly faster inference speeds compared to deep learning-based alternatives. Our approach demonstrated robustness across English and Turkish, indicating its utility in multilingual settings and scenarios with limited computational resources. Thanks to the speed and accuracy of our system, our model can be used to mine large amounts of data that can be used to facilitate the development of new discourse-annotated corpora or as the training data of discourse-focused language models. 

% We plan to train a discourse-aware language model in the future and complete other tasks (argument span classification, sense detection, implicit discourse connective detection) in the discourse parsing pipeline via this language model.

\section{Bibliographical References}\label{sec:reference}

\bibliographystyle{lrec-coling2024-natbib}
\bibliography{lrec-coling2024-example}

\end{document}